\documentclass[lettersize,journal,conference]{IEEEtran}

\usepackage{amsmath,amsfonts}
\usepackage{algorithmic}
\usepackage{algorithm}
\usepackage{array}
\usepackage[caption=false,font=normalsize,labelfont=sf,textfont=sf]{subfig}
\usepackage{textcomp}
\usepackage{stfloats}
\usepackage{url}
\usepackage{verbatim}
\usepackage{xcolor}
\usepackage{graphicx}
\usepackage{caption}
\usepackage{cite}
\usepackage{enumitem}
\usepackage{hyperref}
\usepackage{tabularx}
\usepackage{makecell}
\usepackage{adjustbox}
\usepackage{booktabs}
\usepackage{multirow}
\usepackage{graphicx}

\begin{document}

\title{ Integrating Object Detection, LiDAR-Enhanced Depth Estimation, and Segmentation Models for Railway Environments}

\author{
\IEEEauthorblockN{Enrico F. Giannico\IEEEauthorrefmark{1}, Federico Nesti\IEEEauthorrefmark{1}, Gianluca D'Amico\IEEEauthorrefmark{1}, Mauro Marinoni\IEEEauthorrefmark{1}, \\Edoardo Carosio\IEEEauthorrefmark{1}, Filippo Salotti\IEEEauthorrefmark{2}, Salvatore Sabina\IEEEauthorrefmark{1}, Giorgio Buttazzo\IEEEauthorrefmark{1}}
    \\[1ex]
    \IEEEauthorblockA{\IEEEauthorrefmark{1}\small Department of Excellence in Robotics \& AI, Scuola Superiore Sant'Anna, Pisa, Italy\\
    \IEEEauthorrefmark{2}Progress Rail Signaling S.p.A., Serravalle Pistoiese (PT), Italy}}

\markboth{Journal of \LaTeX\ Class Files,~Vol.~14, No.~8, August~2021}%
{Shell \MakeLowercase{\textit{et al.}}: A Sample Article Using IEEEtran.cls for IEEE Journals}

\IEEEpubid{0000--0000/00\$00.00~\copyright~2021 IEEE}

\maketitle

\begin{abstract}
Obstacle detection in railway environments is crucial
for ensuring safety. However, very few studies address the
problem using a complete, modular, and flexible system that can
both detect objects in the scene and estimate their distance from
the vehicle. Most works focus solely on detection, others attempt
to identify the track, and only a few estimate obstacle distances.
Additionally, evaluating these systems is challenging due to the
lack of ground truth data. In this paper, we propose a modular
and flexible framework that identifies the rail track, detects
potential obstacles, and estimates their distance by integrating three neural networks for object detection, track segmentation, and monocular depth estimation with LiDAR point clouds. To enable a reliable and 
quantitative evaluation, the proposed framework is assessed using a synthetic dataset (SynDRA), which provides accurate ground truth annotations, 
allowing for 
direct performance comparison with existing 
methods. 

The proposed system achieves a
mean absolute error (MAE) as low as 0.63 meters by integrating monocular 
depth maps with LiDAR, enabling not only accurate distance 
estimates but also spatial perception of the scene.

\end{abstract}


\section{Introduction}

\begin{figure}
    \centering
    \includegraphics[width=\linewidth]{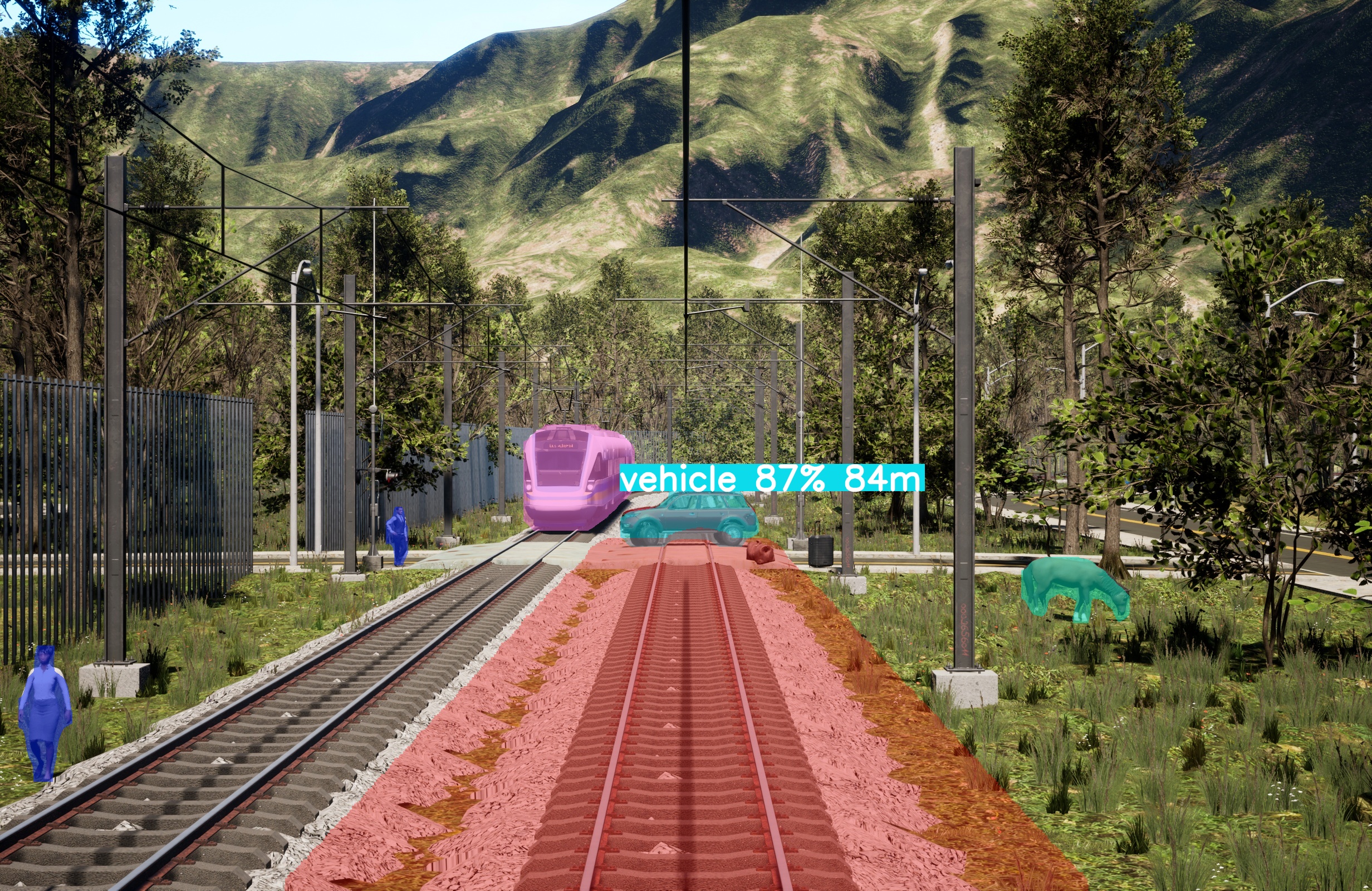}
    \caption{Example of the visual information produced by the proposed framework. Each detected object is shown with a colored segmentation mask, where colors indicate object classes. Objects intersecting the expanded railway track mask, namely potential obstacles, are additionally annotated with their semantic class, detection score, and estimated distance.}
    \label{fig:output_result}
\end{figure}

In the last decade, advancements in artificial intelligence (AI), deep learning, and sensor technology have revolutionized transportation systems, enabling environment perception and higher levels of automation and safety.
However, most advancements in the literature focus on autonomous driving rather than railway systems, where strict regulations and expensive procedures for sensor mounting and data acquisition significantly hinder the development of AI-based solutions.

As reported in the survey by Risti{\'c}-Durrant et al.~\cite{ristic2021review}, although many data acquisition campaigns have been conducted, most of the resulting datasets are not publicly available. Due to the lack of real-world data, the use of photorealistic simulators to generate 
datasets has recently attracted increasing interest. For example, D'Amico et al.~\cite{d2023trainsim} proposed a railway simulator for data generation, while other works~\cite{d2025syndra, degordoa2023scenario, toprak2020conditional, diaz2025towards, broekman2021railenv, neri2022object} focus on 
synthetic datasets for training and evaluation.

Latest Eurostat accident data~\cite{eurostat_rail_safety} suggest collisions are mainly linked to late detection rather than infrastructure or rolling‑stock failures. On‑board obstacle detection systems therefore address a critical residual safety gap, complementing existing protection measures by enabling earlier perception and reaction at train level.

On-board obstacle detection for railway systems presents challenges related to the kinematic and dynamic constraints of the train. At high speed, the braking distance of a train often exceeds the visibility range of the sensors, meaning that obstacles may be detected but not avoided. 
Nonetheless, in low-speed scenarios (such as stations, shunting yards, or level crossings), on-board obstacle detection is a crucial task that increases safety by warning the human operator of any dangerous situation. 
Additional challenges include managing occlusions due to rail curves, low-visibility conditions, detection of small or object categories not present in the training set, and the characterization of false positives and false negatives.

Recent literature on railway obstacle perception addresses rail-track identification, object detection, and distance estimation through separate or loosely integrated solutions, typically relying on camera-based deep learning methods, often supported by additional sensors; however, only a few studies integrate all tasks within railway-specific scenarios, with most research focusing on generic environments.

Overall, existing methods exhibit recurring limitations, such as (i) a lack of modular, flexible systems tailored to railway scenarios, (ii) tasks addressed in isolation rather than within a unified framework, and (iii) distance estimates often derived from bounding-box regions that include background clutter. Moreover, the absence of publicly available datasets with reliable distance ground truth hampers reproducible evaluation and fair comparison.

Motivated by these limitations, this paper proposes a modular obstacle detection framework that integrates the three tasks within a unified pipeline capable of accurately estimating absolute distances under realistic conditions. 
The system is designed to be agnostic of specific neural network architectures, allowing the integration of different object detection and monocular depth estimation backbones.  
To address the scarcity of annotated railway datasets, realistic synthetic data are used for training and fine-tuning, in both normal and critical environmental conditions.
Moreover, the proposed method exploits multi-object tracking and filtering to enhance the stability of distance estimation across consecutive frames. 

Figure~\ref{fig:output_result} illustrates the output of the proposed framework, which generates colored segmentation masks indicating semantic class and detection/distance information for obstacles.

\textbf{Paper contributions}
In summary, this work provides the following contributions:

\begin{itemize}
    \item It proposes a modular and flexible multi-sensor obstacle detection framework;
    \item It fine-tunes state-of-the-art segmentation, object detection, and monocular depth estimation models for railway environments;
    \item It proposes and evaluates 
    data fusion algorithms integrating object detection, depth estimation, and semantic segmentation; 
    \item It exploits multi-object tracking to filter distance estimations across frames and enhance distance prediction accuracy; 
    \item It evaluates the framework on synthetic and augmented data.       
\end{itemize}

\textbf{Paper structure} The remainder of this paper is organized as follows: Section~\ref{s:related} discusses the related literature, Section~\ref{s:method} describes the proposed method, Section~\ref{s:exp} presents the experimental results, and Section~\ref{s:conclusions} discusses the limitations of the proposed approach and future research directions.  
\section{Background and Related Works}
\label{s:related}

This section discusses the state of the art for deep-learning-based methods. 
It then outlines the open challenges and summarizes how they are addressed in the present work. 

Early railway obstacle detection methods decomposed the problem into three independent sub-tasks: rail-track identification~\cite{5625015,qi2013efficient}, object detection~\cite{rodriguez2012obstacle,nakasone2017frontal}, and distance estimation~\cite{kudinov2020perspective,mockel2003multi}.

Recent research replaced traditional methods with neural network models for object detection and semantic segmentation, often combined with depth sensors for distance estimation.
Although few works addressed distance estimation, most of them still rely on specialized distance sensors such as LiDAR or radar. 
An alternative approach is the use of neural networks for monocular distance estimation.

A popular line of work relies on \textit{single camera} approaches to obtain detection and distance estimation. The image is processed by one or more neural networks that generate both outputs. Two kinds of integration are possible: \textit{monolithic}, where a single neural network is employed,
and \textit{modular}, where dedicated object detection and depth estimation networks produce their outputs, which are then merged to obtain a single final result. 

Monolithic integration is provided by Vajgl et al.~\cite{vajgl2022dist} and Azurmendi et al.~\cite{azurmendi2023simultaneous}. The YOLO~\cite{redmon2016you} detector is modified to process captured images for both object detection and distance estimation by extending the prediction vector to include distance information. This approach is easy to implement and avoids the need for additional sensors, but requires ground truth distance annotations, which are typically unavailable in public railway datasets.
Alternatively, Alhasanat et al.~\cite{alhasanat2021retinanet} used RetinaNet~\cite{lin2017focal} as a detector and relied on the known size of a reference object to infer distances of other objects in the scene, which reduces the flexibility of the approach. 

Haseeb et al.~\cite{haseeb2018disnet} adopted a modular integration strategy proposing DisNet. Rather than requiring a priori knowledge about the scene, they used a neural network to exploit the outputs of YOLO, in particular the characteristics of the object bounding boxes, to infer object distance. 
Chen et al.~\cite{chen2019real}, instead, combined an object detection network (YOLOv3~\cite{farhadi2018yolov3} or SSD~\cite{liu2016ssd}) with a monocular depth estimation network based on Monodepth~\cite{godard2017unsupervised}.  
Their approach extracts the bounding boxes of detected objects and computes the histogram of depth values within each region, estimating object distance from the dominant depth interval.
However, this work focuses on estimating the direction of motion and speed of the detected objects rather than accurately determining their distance. 

Similarly, Masoumian et al.~\cite{masoumian2021absolute} employed YOLOv5~\cite{ultralytics_yolo} to detect objects within the scene, while a relative depth map is estimated in parallel using DepthNet~\cite{kumar2018depthnet}. However, it provides only relative measurements, which cannot be directly used in real-word applications.  
To address this limitation, the authors converted relative depth estimates into absolute measurements using object-level information extracted during detection, combined with structural cues from the relative depth map. 
Faseeh et al.~\cite{faseeh2024deep} used a fine-tuned version of YOLOv11~\cite{ultralytics_yolo} to detect objects and a depth encoder–decoder network to generate the depth map. This second network is enhanced with a recurrent convolutional LSTM to process frame sequences
and better capture the spatial and temporal dependencies, improving temporal coherence.
        
Unlike the single-camera approach, the \textit{multi-sensor} approach leverages multiple technologies to enhance accuracy.

A common strategy is to integrate image-based object detection with distance estimates from multi-camera setups, either by selecting the closest point within the object's bounding box~\cite{nair2018camera}, or by stereo-matching bounding boxes from independently processed stereo pairs~\cite{strbac2020yolo, hamad2024object}. 
            
Another possibility is the integration of a millimeter-wave radar and vision sensors. For instance, Wu et al.~\cite{wu2025multi} achieved high-precision detection and tracking of moving objects through feature-level fusion; however, they did not provide an explicit distance estimate for the detected objects. 
            
As an alternative, object detection results can be integrated with LiDAR point clouds~\cite{gao2021multi,zhang2023automatic}.
In these approaches, a semantic segmentation network identified the railway track, which was then used to isolate the relevant subset of the point cloud.  
Reported results showed that such integration improved both detection and distance estimation accuracy. However, rectangular bounding boxes may not accurately delimit object extent and can introduce inaccuracies in distance computation. A similar approach~\cite{favelli2024sensor} has also been applied in the automotive domain.

An overview of the characteristics of the previously mentioned approaches is summarized in Table~\ref{tab:overview}.

\textbf{Open challenges and gaps filled by this paper.}
The works presented above present some difficulties that limit their applicability and scalability in real-world railway scenarios.

Monolithic methods require reference objects or training sets with annotated bounding boxes and distances, limiting adaptability and increasing annotation effort. Modular approaches struggle to convert relative depth estimates into absolute distances. Multi-camera approaches need precise image recalibration, and stereo-based distance accuracy decreases with distance.
Common issues also emerge across most analyzed works: (i) there is a lack of a unified, modular, and flexible system capable of jointly addressing object detection, distance estimation, and railway track identification; (ii) many approaches rely on object localization through bounding boxes, which can include background pixels in depth calculations, reducing distance estimation accuracy; and (iii) the absence of publicly available datasets with reliable distance ground truth further complicates the development of robust solutions, preventing reproducible performance evaluations.

The proposed approach addresses these gaps by establishing a modular and flexible framework that integrates railway track segmentation, object detection, and monocular depth estimation with LiDAR measurements.  
Absolute depth information is produced directly within the system, improving flexibility and generalization, while distance estimation relies on depth values extracted from segmentation masks, preventing the inclusion of background pixels.
Multi-object tracking further enhances the temporal stability of distance estimates.  
To the best of our knowledge, this is the first work 
evaluated using realistic synthetic datasets, addressing the lack of freely available real-world data with ground-truth, particularly in railway scenarios.

\begin{table}[t]
\centering
\caption{Overview of existing approaches for integrating object detection and distance estimation. 
Loc. denotes the localization method. 
Dist. Est. indicates the distance estimation technique. 
Track Det. indicates whether railway track detection is included.}
\label{tab:overview}
\scriptsize
\renewcommand{\arraystretch}{0.5}
\setlength{\tabcolsep}{2.9pt}

\begin{tabular}{l l c c c c c}
\toprule
\textbf{Type} & \textbf{Ref.} & \textbf{Scope} & \textbf{Detector} & \textbf{Loc.} & \textbf{Dist. Est.} & \textbf{Track Det.} \\
\midrule

\multicolumn{7}{l}{\textbf{Single camera}} \\
\cmidrule(lr){1-7}
\multirow{3}{*}{Monolithic}
& \cite{vajgl2022dist} & General & YOLOv3 & BBox & Custom & No \\
& \cite{azurmendi2023simultaneous} & Custom & YOLOv5 & BBox & Custom & No \\
& \cite{alhasanat2021retinanet} & General & RetinaNet  & BBox & Ref. obj. & No \\
\cmidrule(lr){1-7}
\multirow{5}{*}{Modular}
& \cite{haseeb2018disnet} & Rway/Auto & YOLOv1 &  BBox & DisNet & No \\
& \cite{chen2019real} & Rway/Auto & YOLOv3/SSD  & BBox & MonoDepth & No \\
& \cite{masoumian2021absolute} & Short dist. & YOLOv5  & BBox & DepthNet & No \\
& \cite{faseeh2024deep} & General & YOLOv11 & BBox & Custom & No \\
\midrule

\multicolumn{7}{l}{\textbf{Multi-sensor}} \\
\cmidrule(lr){1-7}
\multirow{3}{*}{Multi-camera}
& \cite{nair2018camera} & Short dist. & MobileNet-SSD & BBox & Stereo & No \\
& \cite{strbac2020yolo} & Short dist. & YOLOv3 & BBox & Stereo & No \\
& \cite{hamad2024object} & Short dist. & MobileNet-SSD & BBox & Stereo & No \\
\cmidrule(lr){1-7}
Cam+Radar & \cite{wu2025multi} & Rway & YOLOv8 & BBox & Radar & No \\
\cmidrule(lr){1-7}
\multirow{8}{*}{Cam+LiDAR} & \cite{gao2021multi} & Rway & SSD & BBox & LiDAR & Yes \\
& \cite{zhang2023automatic} & Rway & SSD & BBox & LiDAR & Yes \\
& \cite{favelli2024sensor} & Auto & YOLOv7 & BBox & LiDAR & No \\
\cmidrule(lr){2-7}
& \textbf{Our} & \textbf{Rway} & \textbf{YOLOv8/11} &  \textbf{Mask} & \makecell[c]{\textbf{LiDAR} \\ \textbf{+ MiDaS}} & \textbf{Yes} \\
\bottomrule
\end{tabular}
\end{table}

\section{Proposed approach}
\label{s:method}
This section describes the proposed approach. Section~\ref{s:method_proposed} presents the overall architecture; Section~\ref{s:method_nn} details the neural networks module; Section~\ref{s:method_fusion} describes the distance estimation module, and Section~\ref{s:method_viz} shows the output visualization module.
\begin{figure*}[h]
    \centering
    \includegraphics[width=\linewidth]{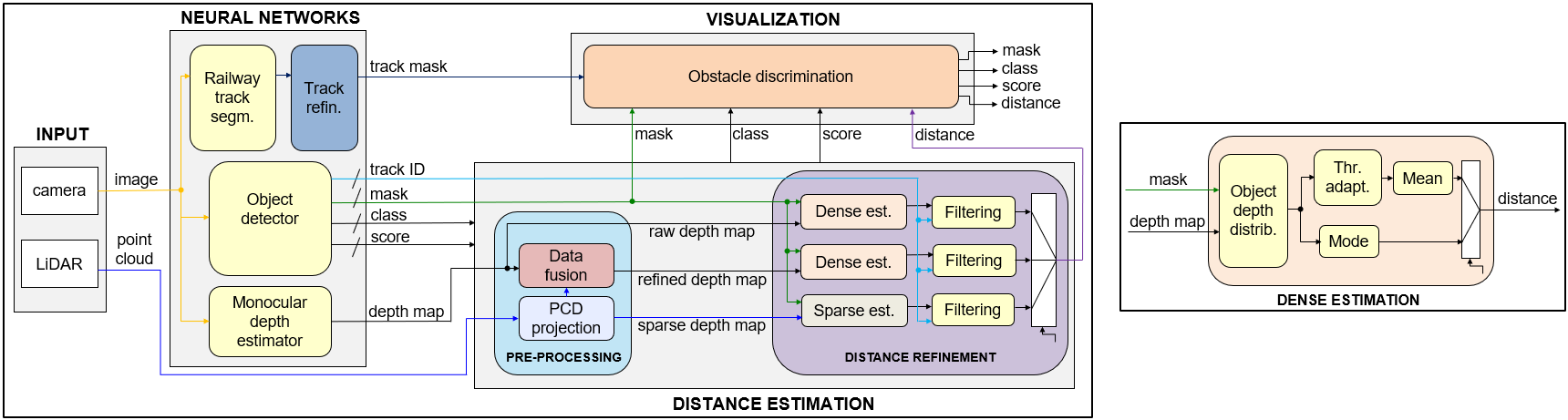}
    \caption{Block diagram of the proposed architecture. A detailed view of the Dense estimation block is shown on the right.}
    \label{fig:architecture}
\end{figure*}
\subsection{Proposed Architecture}\label{s:method_proposed}
The proposed architecture is illustrated in
Figure~\ref{fig:architecture} and consists of four main modules described below. 

\textbf{Input}: This module is responsible for acquiring data from the external environment. The inputs include a camera, and one or more LiDAR sensors. 
The camera captures the frontal view of the train, providing the primary input for obstacle detection and coarse distance estimation. The LiDAR point cloud is used to obtain accurate distance measurements.

\textbf{Neural networks}: This module includes three models, each addressing a specific sub-task: Railway track segmentation, Object detection, and Monocular depth estimation. Each network processes the same input frame acquired by the camera.

The Railway track segmentation network generates a mask of the track, enabling its precise localization and focusing the analysis on potential obstacles located on or near the track, rather than on all objects present in the scene. 
The Object detector identifies objects in the image and outputs, for each instance, the class label, confidence score, and spatial location, represented by a segmentation mask.
Optionally, it can also provide a tracking ID for each object. The Monocular depth estimator produces a dense depth map in which each pixel encodes the estimated absolute distance from the camera.
The outputs of these networks are then post-processed and fused, as detailed in Sections~\ref{s:method_fusion} and~\ref{s:method_viz}.

\textbf{Distance estimation}: It leverages the detected object information, the generated depth map, and the LiDAR point clouds to estimate the distance of each object from the sensors, according to the selected estimation strategy.

\textbf{Visualization}: It visualize results and can be used as a driver-assistance system in the train cabin. Namely, it highlights objects on or near the railway track ahead of the vehicle, distinguishing between non-hazardous items and potentially dangerous obstacles based on their distance from the track. 

This information can provide early warnings, leaving the train operator the decision to stop, slow down, or ignore the alert.
\subsection{Neural Networks}\label{s:method_nn}

\begin{figure}[!ht]
    \centering    
    \includegraphics[width=0.80\columnwidth]{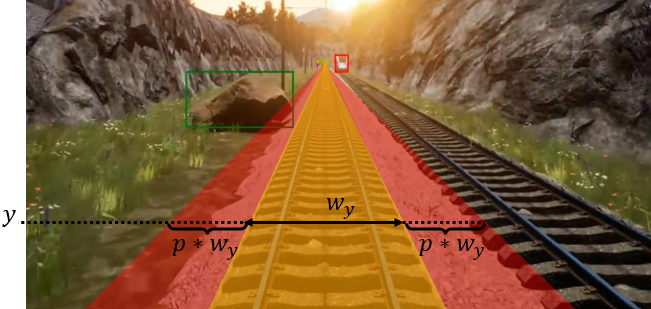}
    \caption{Example of a refined railway track segmentation mask.}
    \label{fig:roi_refinement}
\end{figure}

\textbf{Railway track segmentation:}
 This network processes input images from the camera to generate a segmentation mask of the railway track, distinguishing between ``track" and ``no-track" regions. 
 To define an accurate region of interest, the track mask is expanded horizontally by a factor $p>0$ to include nearby objects that, while not directly on the track, may pose potential hazards. At each image coordinate $y$, the mask width $w_y$ is expanded by a factor $p w_y$ on both sides, resulting in a final width of $(1+2p)w_y$. When obstacles are directly on the track, the segmentation model may produce incomplete or fragmented masks, likely because such cases were underrepresented in the training data. To address this problem, inaccurate masks are replaced with the last valid mask. Since the track appearance is similar between consecutive frames, this strategy provides temporally consistent masks. A reduced accuracy has been observed in the presence of sharp curves or steep slopes, but such scenarios are rare, since railway lines typically have large curvature radii.
The result of the refinement process is illustrated in Figure~\ref{fig:roi_refinement}.

\textbf{Object detection:}
The Object detector processes the input image and outputs a list of detected objects, each described by its estimated semantic class, confidence score, and bounding box. Alternatively, advanced networks can output more refined segmentation masks rather than rectangular bounding boxes. The proposed architecture is flexible enough to allow the use of both bounding boxes and segmentation masks. However, segmentation masks enable a clearer interpretation of results, and, most importantly, a more accurate distance estimation for each object, as explained in Section~\ref{s:method_fusion}.
Common pre-trained object detectors might require fine-tuning and semantic class re-organization, as explained in Section~\ref{s:exp_setup}.

\textbf{Monocular depth estimation:}
The Monocular depth estimator processes the input image and outputs a dense depth map with the same dimensions as the input.
Such models typically suffer from generalization issues, which originate from two key factors: (i) monocular depth estimation is an ill-posed task, as the depth scale factor is lost in the projection, making the output often relative rather than absolute and thus not directly usable in real applications;
and (ii) these networks are usually trained on custom datasets characterized by specific input distributions and output representations. In fact, ground truth depth maps are typically generated using LiDAR, stereo cameras, or RGB-D sensors. 
Hence, a dedicated fine-tuning process is required to obtain a network capable of correctly estimating depth for the highly-structured data typical of railway environments.
\subsection{Distance Estimation}\label{s:method_fusion}
The distance estimation module takes as input the detection information produced by the Object detector, as well as the depth information provided by the Monocular depth estimator and the LiDAR sensor.
Within the module, this information is first pre-processed and then forwarded to the Distance refinement block.

    \textbf{1) Pre-processing:}    
    As a preliminary step, the LiDAR point cloud data (PCD) are pre-processed for use in the Distance refinement stage. Specifically, the three-dimensional LiDAR point cloud is projected  onto the camera image plane by means of the PCD projection module, in order to establish a correspondence between LiDAR measurements and image pixels, resulting in a two-dimensional \textit{sparse depth map}.
    The projection is performed by standard pinhole camera model that uses extrinsic and intrinsic matrices~\cite{Hartley2004}. 

    The dense depth map produced by the fine-tuned monocular network is processed in two ways. On the one hand, it is directly forwarded to the Distance refinement block (\textit{raw depth map}); on the other hand, it is refined within the Data fusion block by linearly interpolating corrections derived from the projected LiDAR point cloud. Linear interpolation is adopted due to its low computational cost compared to more complex methods. To this end, the projected LiDAR points are used to compute local depth residuals with respect to the monocular prediction. These residuals are then interpolated over the image plane to obtain a dense correction field, which is added to the original raw depth map to produce a refined estimate.
    Let $D_m(x,y)$ denote the raw monocular depth map and $\{(x_i, y_i, d_i)\}_{i=1}^{N}$ the set of LiDAR points projected onto the image plane.
    For each LiDAR projected point at pixel coordinates $(x_i,y_i)$, a depth residual is computed as
    \begin{equation}
    r_i(x_i, y_i) = d_i(x_i, y_i) - D_m(x_i, y_i).
    \end{equation}
    The sparse residuals are linearly interpolated over the image domain to obtain a dense residual map $R(x,y)$. The \textit{refined depth map} $D_r(x,y)$ is obtained by adding $R(x,y)$ to $D_m(x,y)$, and the result is forwarded to the Distance refinement block.
    
    In this way, the Distance refinement block leverages three types of depth inputs: the \textit{sparse depth map}, the \textit{raw dense depth map}, and the \textit{refined dense depth map}. 
    
    \textbf{2) Distance refinement:}
    The detected object segmentation mask is used to compute the object distance estimate using alternative or combined strategies.
    
    \textbf{Sparse estimation:} When using the \textit{sparse depth map}, the distance is estimated by computing the mode of the depth distribution within the segmentation mask. Let $\Omega$ denote the sparse depth map, $\mathcal{R} \subset \Omega$ the segmentation mask of the detected object, and $p_{\mathcal{R}}(d)$ the distribution of depth values within $\mathcal{R}$. The object distance is then given by the mode of this distribution.

\begin{figure}[!ht]
    \centering    
    \includegraphics[width=\columnwidth]{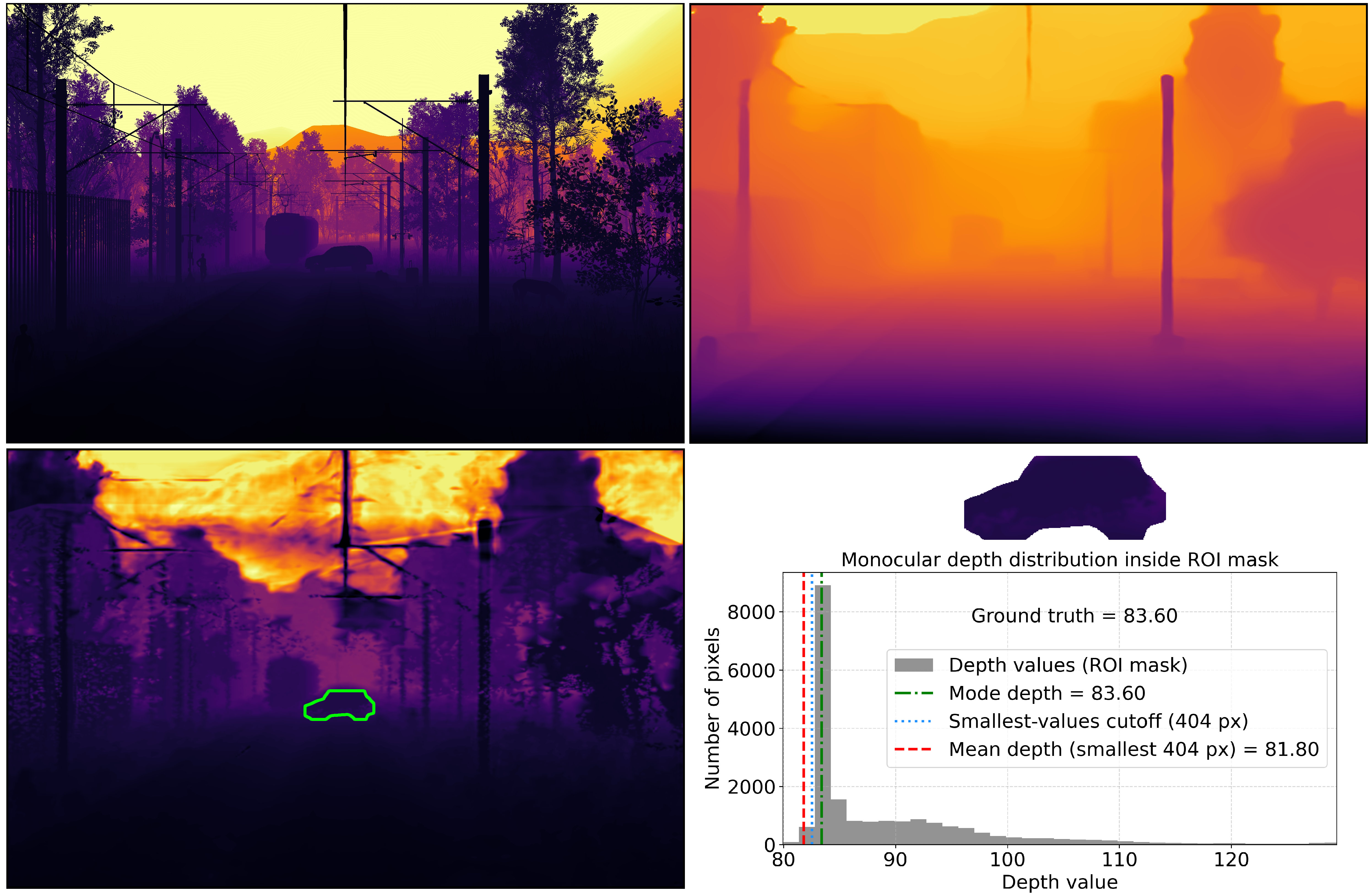}
    \caption{Top-left: ground-truth depth map of the scene illustrated in Figure 1. Top-right: corresponding monocular depth map before fine-tuning. Bottom-left: monocular depth map after fine-tuning and LiDAR refinement. Bottom-right: distance estimation obtained from the obstacle segmentation mask.}
    \label{fig:depth_estim}
    \end{figure}

    \textbf{Dense estimation:} When using a dense depth map, two strategies can be adopted: either using the mode, as in the sparse approach, or employing a more conservative strategy. In the latter case, the mean depth is computed over a subset of depth values determined by the size of the segmentation mask. Let $\mathcal{R}' \subset \mathcal{R}$ denote the subset containing the $k\%$ smallest depth values within $\mathcal{R}$. The distance estimate obtained via Threshold adaptation is then computed as the mean over this subset. Figure~\ref{fig:depth_estim} shows the depth maps before fine-tuning and after fine-tuning followed by LiDAR refinement, highlighting its similarity to the ground-truth depth map. The figure also illustrates object distance estimation, computed using either the mode of the depth distribution or the mean of the k\% smallest depth values within the segmentation mask.
    
    \textbf{Temporal filtering:}    
    In addition to per-frame estimation, the tracking ID produced by the detector is used to incorporate temporal information, filtering out outliers and improving the stability of distance estimates across frames. For each tracked object, a sliding temporal window of consecutive filtered distance estimates associated with the same tracking ID is maintained. A weighted mean is then computed, assigning higher weights to more recent estimates. This aggregation reduces short-term fluctuations in the estimated object distance. If tracking is unavailable, the Filtering block is bypassed and the per-frame distance estimate is used directly.
    
    Depending on the chosen strategy, the corresponding distance estimate can be forwarded to the Obstacle discrimination block.

\subsection{Visualization}\label{s:method_viz}

The visualization module presents visual information about the perceived scene to assist the conductors by alerting them when a potential obstacle is detected. Only objects whose segmentation masks intersect the refined railway track mask are considered obstacles. All detected objects are shown with colored segmentation masks according to their semantic class. For obstacles, the semantic class, detection confidence, and estimated distance from the train are also displayed, while bounding boxes are omitted to avoid cluttering the scene. An example of the output is shown in Figure~\ref{fig:output_result}.

\section{Experimental Results}
\label{s:exp}

This section presents the results of the experiments carried out to evaluate the performance of the proposed approach. Section~\ref{s:exp_setup} describes the system setup, including datasets, the LiDAR point cloud noise model, the implemented neural networks, the fine-tuning procedures, and their integration into a unified system. Section~\ref{s:exp_eval} focuses on evaluating the performance of the fine-tuned models and the complete system.

\subsection{Experimental Setup}\label{s:exp_setup}
\subsubsection{Hardware and software details}
The experiments were conducted offline on a server equipped with an NVIDIA A100 GPU, used to train and evaluate the integrated system. From a software perspective, the operating system was Ubuntu 22.04.5 LTS (Jammy Jellyfish), and all neural networks were implemented in PyTorch within a Python environment (Python 3.10.16, PyTorch 2.6.0, CUDA 12.4).

\subsubsection{Datasets}
Several datasets were used in this work:

\begin{itemize}
    \item RailSem19~\cite{zendel2019railsem19}: a publicly available dataset for semantic segmentation in railway scenarios. Similarly to previous work~\cite{d2025syndra}, the dataset was customized by generating dedicated ground-truth labels for binary semantic segmentation. This dataset was used to fine-tune the rail segmentation network.
    \item OSDaR23~\cite{tagiew2023osdar23}: a multi-modal, publicly available dataset for scene perception in railway environment. Similarly to previous work~\cite{d2025syndra}, the original geometric annotations of the main rail track were converted to binary segmentation masks. This dataset was used to evaluate the performance of the fine-tuned rail segmentation network.
    \item COCO~\cite{lin2014microsoft}: most of its standard 80 object classes are not relevant to the railway domain. For this reason, a class remapping strategy was defined to map relevant original classes into five super-classes of interest: \textit{person}, \textit{vehicle}, \textit{animal}, \textit{tool}, and \textit{train}. The remaining classes were grouped into a new class called \textit{other}. This dataset was used to fine-tune and evaluate the Object detector.
    \item OSDaR-AR~\cite{nesti2026osdararenhancingrailwayperception}: an augmented version of OSDaR23 in which external objects are inserted into RGB frames. The subsequences extracted from 3\_fire\_site\_3.1 and 6\_station\_klein\_flottbek\_6.2 that include a \textit{cow}, an \textit{elephant}, a \textit{horse}, and a \textit{pedestrian} were considered, for a total of 8 sequences with 100 frames each. This dataset was used to evaluate the detection performance of the overall system, but ignoring the distance estimation module for lack of dense depth ground truth maps.
    \item SynDRA~\cite{d2025syndra}: a family of synthetic datasets for railway applications generated using a custom simulator developed in Unreal Engine 5. One data split, referred to as the ``depth split'', containing 1339 frames of RGB images and dense depth maps, was employed to fine-tune and evaluate the monocular depth estimation network. Another split, referred to as the ``evaluation split'', includes four synthetic sequences covering complex, safety-critical railway scenarios underrepresented in public datasets: (i) a \textit{rural} open-field railway alongside roads and natural terrain; (ii) an \textit{urban} area near buildings with streets, vehicles, and pedestrians; (iii) a \textit{railway station} with multiple tracks, platforms, workers, and luggage; and (iv) a \textit{shunting yard} with trains, workers, vehicles, and industrial facilities. 
    Each environment contains $\approx$2000 frames with RGB images, depth maps, LiDAR point clouds, and annotations for semantic segmentation and 2D/3D bounding boxes. This dataset was used for the quantitative evaluation of the overall system.
\end{itemize}

\subsubsection{LiDAR point cloud degradation model}

A single point cloud per frame was obtained by merging measurements from three LiDAR sensors mounted at different positions (left, center, and right) in the SynDRA~\cite{d2025syndra} dataset. To simulate realistic measurements uncertainty, depth noise and dropout effects were applied, with both increasing for points at larger distances. This approximates the range-dependent behavior of typical mid-range spinning automotive LiDAR sensors. A fixed random seed ensures that the generated perturbations are reproducible while varying across frames.  

\subsubsection{Neural network models and fine-tuning}
The architecture in Section~\ref{s:method} acts as a general blueprint to set up an obstacle detection module for railway systems, allowing the use of different models. In this work, experimental validation was conducted using a specific combination of neural networks. 

\textbf{Railway track segmentation:}
The network selected for Railway track segmentation is DDRNet23-Slim~\cite{pan2022deep}, a lightweight dual-resolution architecture designed for efficient real-time semantic segmentation. Its excellent performance for this task was demonstrated in previous work~\cite{d2025syndra}, so the same setup was adopted. The publicly available model, pre-trained on the Cityscapes dataset~\cite{7780719} with 19 classes, was modified to output two classes: “track” and “no-track”. Fine-tuning was performed on the tram-less, binary segmentation version of RailSem19~\cite{zendel2019railsem19}.  
The network was trained for 30 epochs using the Adam optimizer with PyTorch default settings and a learning rate of 0.001.

\textbf{Object Detection:}
The Ultralytics YOLO framework ~\cite{ultralytics_yolo} was selected as the Object detector, due to its high flexibility, ease of use, and strong real-time performance. The publicly available weights of the considered YOLO variants (v5n/m/x, v8n/m/x, and v11n/m/x), including both detection and segmentation models, were pre-trained on the COCO~\cite{lin2014microsoft} dataset, whose remapped version was then used to fine-tune the classification head, as described in the previous section.
After five training epochs, using a learning rate of 0.01 and an input image resolution of 640×640, the fine-tuned models were able to detect the selected object super-classes. The availability of multiple architectures and model variants provides a high degree of flexibility, allowing the selection of the most suitable configuration according to real-time constraints, such as target frame rate and accuracy requirements, making the overall system highly adaptable. Among the evaluated models, the YOLOv11x segmentation variant, used with the default BoT-SORT tracker, was selected because it achieved the highest accuracy and enabled more precise object localization through pixel-level contour prediction.

\textbf{Monocular depth estimation:}
Due to its strong generalization capability, MiDaS~\cite{ranftl2020towards} was selected as Monocular depth estimator. The latest version (v3.1) with a Swin2L-384 backbone was adopted for its balance between accuracy and inference time. However, like many monocular depth models, MiDaS produces relative depth estimates that represent depth relationships within the scene rather than absolute distances.
To address this limitation, the model was fine-tuned using the SynDRA~\cite{d2025syndra} ``depth split'' introduced in the previous subsection. To the best of our knowledge, SynDRA is the only public dataset providing dense depth data for railway environments. Since standard mean squared error (MSE) loss performs poorly over the full considered range ([0, 655] m), the network was fine-tuned using three weighted MSE variants.

\vspace{1cm}
Most detectable and avoidable objects lie within 200 m. To emphasize this critical range, the first variant (\textit{threshold-based weighting}) assigns full weight to depth values below \(T = 200\) m and a reduced weight \(\alpha = 0.1\) to values beyond \(T\), focusing on shorter distances while retaining some contribution from farther regions.
The second variant, (\textit{linear decay}) emphasizes closer depths with full weight below \(T\), linearly decreasing to \(\alpha\) at the maximum depth \(D_{\max} = 655\) m.
To mitigate depth distribution imbalances, a \textit{frequency-based weighting} scheme was also introduced. The depth range is partitioned into 22 intervals of approximately 30 m. Each interval is assigned a weight inversely proportional to its average pixel frequency, computed over the 1339 frames of the SynDRA ``depth split''.

All three strategies are modeled as the following weighted MSE loss:
\begin{equation}
\mathcal{L} = \frac{1}{N} \sum_{i=1}^N w_i \left(\hat{d}_i - d_i\right)^2,
\end{equation}
where \(w_i\) denotes the pixel weight computed according to one of the strategies, \(N\) is the total number of pixels, \(\hat{d}_i\) the predicted depth, and \(d_i\) the ground truth.
The Adam optimizer was used with a learning rate of 0.001, and the mean absolute error (MAE) was used as the evaluation metric.

This approach yielded three models capable of producing absolute distance estimates, each optimized for a specific range: short ([0, 200] m), medium ([200, 300] m), and long ([300, 655] m). Among the obtained models, the fine-tuned MiDaS model with the threshold-based weighting strategy was
selected, as it improves performance in the near-range interval, which is particularly relevant for railway safety applications.

\subsubsection{Networks synchronization}
The framework runs three concurrent threads, one per network. The main YOLO thread acquires the input stream and processes each frame for object detection. Each new frame simultaneously triggers the MiDaS and DDRNet threads, which produce the depth map and track segmentation mask in parallel. The YOLO thread then waits for both processes to complete before merging all outputs, namely detection results, depth estimates, and the segmentation mask, into the final obstacle detection decision. This synchronized pipeline repeats for every input frame.
   
\begin{table}[t]
\centering
\caption{Performance results of the selected fine-tuned models: DDRNet23, YOLOv11, and MiDaS v3.1.}
\label{tab:combined_results}
\scriptsize
\setlength{\tabcolsep}{1pt}
\renewcommand{\arraystretch}{0.8}
\begin{tabular*}{\linewidth}{@{\extracolsep{\fill}}lccc}
\toprule
\multicolumn{4}{c}{\textbf{Fine-tuned DDRNet23 (Slim) -- OSDaR23}} \\
\midrule
Accuracy & IoU & Precision & Recall \\
0.99 & 0.94 & 0.99 & 0.95 \\
\midrule
\multicolumn{4}{c}{\textbf{Fine-tuned YOLOv11 (Segmentation) -- COCO}} \\
\midrule
 & Nano & Medium & Extralarge \\
\midrule
mAP@50      & 0.63 & 0.75 & \textbf{0.78} \\
mAP@50--95  & 0.39 & 0.49 & \textbf{0.51} \\
\midrule
\multicolumn{4}{c}{\textbf{Fine-tuned MiDaS v3.1 (Swin2L-384) -- SynDRA}} \\
\midrule
 & Threshold-based & Linear decay & Frequency-based \\
\midrule
MAE [0--655]   & 41.5 & 35.1 & 37.0 \\
MAE [0--200]   & \textbf{12.2} & 14.7 & 33.0 \\
MAE [200--300] & 46.7 & \textbf{33.7} & 44.8 \\
MAE [300--655] & 94.4 & 73.0 & \textbf{42.8} \\
\bottomrule
\end{tabular*}
\end{table}

\subsection{Results Evaluation}\label{s:exp_eval}
\subsubsection{Evaluation of fine-tuned models}

Table~\ref{tab:combined_results} summarizes the performance of the fine-tuned networks used in this study.  

The fine-tuned DDRNet model showed strong performance on the OSDaR23 dataset. 
The fine-tuned YOLOv11 segmentation models, evaluated on the COCO validation set, performed consistently with the publicly available models achieving better results due to training on a subset of classes representing typical railway objects (see Section~\ref{s:exp_setup}).
Finally, the fine-tuned MiDaS models were evaluated on the test set of the SynDRA ``depth split'' using the different weighted MSE loss variants described in Section~\ref{s:exp_setup}. The overall mean absolute error (MAE) across the full distance range [0, 655] m resulted to be roughly 35-42 m, while the models performed better when optimized for specific ranges: the \textit{threshold-based weighting} excelled at short distances, the \textit{linear decay} strategy at medium distances, and the \textit{frequency-based weighting} at long distances.

\subsubsection{Evaluation of the complete setup}
 To evaluate the system, ground-truth annotations were considered only for objects that were visible for more than 75\% of their extent, had an area exceeding 1200 pixels (at $\approx$2.5k resolution), were located within 200 m of the sensor, and fell inside the designated dangerous area near the track. This filtering ensured reliable annotations and excluded excessively small, heavily occluded, distant, or off-center objects that could bias the evaluation.

\begin{figure}[h] 
    \centering
    \includegraphics[width=\linewidth]{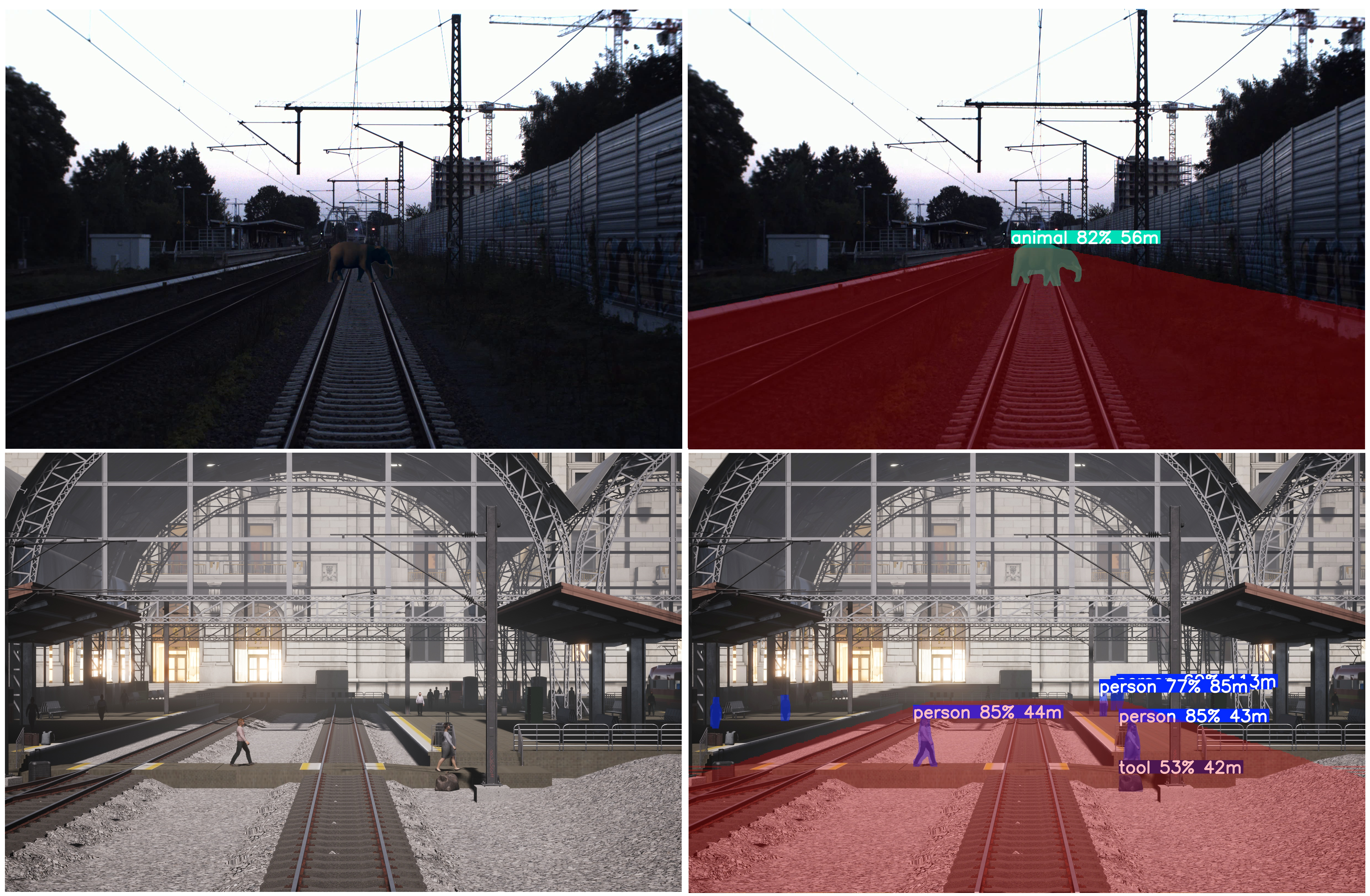}
    \caption{Examples of the system's output on OSDaR-AR (top) and SynDRA (bottom).}
    \label{fig:evaluation_examples}
\end{figure}

Examples of the system’s output are shown in Figure~\ref{fig:evaluation_examples}.

\textbf{OSDaR-AR:} As shown in Table~\ref{tab:qualitative_detection}, all considered objects were detected with a true positive rate (TPR) ranging from 63\% to 99\%, with an IoU between 0.69 and 0.80. An exception was observed for the horse in the 6.2 subsequence, due to the very low contrast between the object and the background, which made detection difficult. The use of YOLO tracking did not significantly affect the results. Performance was higher for larger objects compared to smaller ones; however, in most cases, it remained acceptable. Since no dense ground truth depth maps are provided in the dataset, no distance estimation results are provided. 

\textbf{SynDRA:} For a full quantitative evaluation, the SynDRA ``evaluation split'' described in Section~\ref{s:exp_setup} was used, comprising urban, shunting yard, station, and rural scenarios.  
Detection and distance estimation results are summarized in Tables~\ref{tab:detection} and~\ref{tab:distance}, respectively.

\begin{itemize}
\item \noindent\textbf{Detection accuracy:} Detection performance is generally consistent across scenarios, with slight degradation in the station and shunting yard. The train class achieves the highest detection rate and IoU due to its large, regular geometry. The person class ranks second, though accuracy drops in the station scenario due to occlusions and clutter, while localization remains stable. Vehicles show lower true positive rates than persons but maintain competitive IoU, indicating most errors occur at detection rather than mask refinement. In rural scenario, performance drops further, occasionally misclassifying a bus as a train. Animals are more challenging, with lower detection rates due to high intra-class variability, though IoU is acceptable when detected. Tools, being small and heterogeneous, have the lowest detection rates but achieve relatively high IoU when detected.
Overall, enabling YOLO tracking has minimal impact, with accuracy remaining as without it.

\item \noindent\textbf{Distance estimation accuracy:} Significant differences emerge when estimating depth using LiDAR sparse, MiDaS raw, and MiDaS refined depth maps. LiDAR mode-based measurements remain highly reliable, as using only depth pixels within the YOLO-predicted object mask mitigates outliers and background interference, resulting in low MAE. 
MiDaS raw depth maps exhibit substantially higher errors due to noise and uncertainty, despite providing dense absolute predictions after fine-tuning. Mode-based estimates generally outperform mean-based ones in most scenarios, reflecting skewed and noisy depth distributions; however, in the station scenario, mean-based aggregation yields lower errors. MiDaS refined maps greatly reduce MAE compared to MiDaS raw, approaching LiDAR accuracy while maintaining dense spatial coverage, thus providing a good compromise between accuracy and completeness.

Temporal aggregation affects different estimators in distinct ways. For noise-sensitive statistics such as the mean, increasing the tracking window reduces MAE by smoothing fluctuations. In contrast, when robust statistics (e.g., the mode) are applied to low-noise depth data such as LiDAR or MiDaS refined maps, longer windows may degrade accuracy due to over-smoothing, making shorter spans (1-5 frames) preferable. The optimal window size thus depends on application requirements, estimation strategy, selected metric, and frame rate.

In conclusion, fine-tuning the monocular depth network on the SynDRA dataset provided initial dense absolute depth maps with limited accuracy. LiDAR offered reliable but sparse depth, limiting pixel-level detail. Combining LiDAR and MiDaS via linear interpolation produced dense and reliable depth maps, while temporal filtering based on YOLO tracking smoothed per-object depth estimates over time, improving per-frame stability. 

\begin{table}[t]
\centering
\caption{Detection results on OSDaR-AR.}
\label{tab:qualitative_detection}
\scriptsize
\setlength{\tabcolsep}{1.8pt}
\renewcommand{\arraystretch}{0.8}
\begin{tabular}{lccccc}
\toprule
\textbf{Scenario} & \textbf{Metric} & \textbf{3.1 NoTracking} & \textbf{3.1 Tracking} & \textbf{6.2 NoTracking} & \textbf{6.2 Tracking}\\
\midrule
\multirow{2}{*}{Cow}
& TPR (\%) & 63.0  & 63.0 & 66.0 & 64.0 \\
& IoU@0.5 & 0.78 & 0.78 & 0.80 & 0.80 \\
\midrule
\multirow{2}{*}{Elephant}
& TPR (\%) & 99.0 & 99.0 & 93.0 & 92.0 \\
& IoU@0.5 & 0.78 & 0.78 & 0.79 & 0.79 \\
\midrule
\multirow{2}{*}{Horse}
& TPR (\%) & 94.4 & 94.4 & 30.0 & 30.0 \\
& IoU@0.5 & 0.69 & 0.69 & 0.73 & 0.73 \\
\midrule
\multirow{2}{*}{Pedestrian}
& TPR (\%) & 97.7 & 97.7 & 96.0 & 94.0 \\
& IoU@0.5 & 0.74 & 0.74 & 0.74 & 0.74 \\
\bottomrule
\end{tabular}
\end{table}

\begin{table}[t]
\centering
\caption{Quantitative results: detection performance across SynDRA scenarios. NP: Not Present; ND: Not Detected; NT: No Tracking; T: Tracking.}
\label{tab:detection}
\scriptsize
\setlength{\tabcolsep}{2.3pt}
\renewcommand{\arraystretch}{0.8}
\begin{tabular}{lcccccc}
\toprule
\textbf{Scenario} & \textbf{Metric} & \textbf{Person} & \textbf{Vehicle} & \textbf{Animal} & \textbf{Tool} & \textbf{Train} \\
\midrule
\multirow{3}{*}{Urban}
& TPR (\%) NT/T & 86.7/86.6 & 67/67 & 58.5/58.0 & ND & 98.7/98.7 \\
& IoU@0.5 NT/T & 0.62/0.62 & 0.75/0.75 & 0.67/0.67 & ND & 0.88/0.88 \\
& mIoU@0.5 NT/T & \multicolumn{5}{c}{0.79/0.79} \\
\midrule
\multirow{3}{*}{Rural}
& TPR (\%) NT/T & 88.5/88.3 & 21.7/21.2 & 47.7/46.9 & NP & NP \\
& IoU@0.5 NT/T & 0.60/0.60 & 0.87/0.87 & 0.69/0.69 & NP & NP \\
& mIoU@0.5 NT/T & \multicolumn{5}{c}{0.64/0.64} \\
\midrule
\multirow{3}{*}{Shunting Y.}
& TPR (\%) NT/T & 78.5/77.6 & NP & NP & ND & 98.3/98.1 \\
& IoU@0.5 NT/T & 0.68/0.68 & NP & NP & ND & 0.89/0.89 \\
& mIoU@0.5 NT/T & \multicolumn{5}{c}{0.79/0.79} \\
\midrule
\multirow{3}{*}{Station}
& TPR (\%) NT/T & 72.6/72.5 & NP & NP & 4.9/4.6 & NP \\
& IoU@0.5 NT/T & 0.64/0.64 & NP & NP & 0.83/0.83 & NP \\
& mIoU@0.5 NT/T & \multicolumn{5}{c}{0.66/0.65} \\
\bottomrule
\end{tabular}
\end{table}

\begin{table}[t]
\centering
\caption{Quantitative results: mean absolute error (MAE) of distance estimation (m) for different tracking window sizes. A window size of 1 indicates that only the current distance estimate is used.}
\label{tab:distance}
\scriptsize
\setlength{\tabcolsep}{6pt}  
\renewcommand{\arraystretch}{0.8}
\begin{tabular*}{\linewidth}{@{\extracolsep{\fill}}lccccc}
\toprule
\textbf{Tracking window} & 1 & 5 & 17 & 29 & 51 \\
\midrule
\textbf{Urban} &  &  &  &  &  \\
LiDAR (mode) & \textbf{0.45} & 0.47 & 0.96 & 1.45 & 2.32 \\
RAW (mean k\%) & 16.97 & 16.84 & 16.44 & 16.18 & \textbf{15.67} \\
RAW (mode) & 13.27 & \textbf{12.23} & 12.58 & 16.2 & 16.39 \\
REF (mean k\%) & 5.27 & 5.16 & 5.04 & 4.23 & \textbf{4.14} \\
REF (mode) & \textbf{0.63} & 0.67 & 1.14 & 1.61 & 2.49 \\
\midrule
\textbf{Rural} &  &  &  &  &  \\
LiDAR (mode) & 2.10 & \textbf{0.58} & 1.66 & 2.92 & 5.08 \\
RAW (mean k\%) & 17.95 & 18 & 16.99 & 16.35 & \textbf{15.35} \\
RAW (mode) & 11.98 & 11.03 & 10.94 & \textbf{10.82} & 12.34 \\
REF (mean k\%) & 6.13 & 6.24 & 3.66 & \textbf{3.25} & 3.50 \\
REF (mode) & 1.68 & \textbf{1.15} & 2.19 & 3.17 & 4.95 \\
\midrule
\textbf{Shunting Yard} &  &  &  &  &  \\
LiDAR (mode) & \textbf{3.90} & 3.93 & 4.34 & 4.85 & 5.77 \\
RAW (mean k\%) & 18.05 & 17.82 & 17.46 & 17.14 & \textbf{16.62} \\
RAW (mode) & 12.19 & 11.53 & 11.24 & 11.02 & \textbf{10.83} \\
REF (mean k\%) & 3.11 & 3.24 & 2.93 & \textbf{2.89} & 3.02 \\
REF (mode) & \textbf{3.96} & 4.06 & 4.60 & 5.16 & 6.41 \\
\midrule
\textbf{Station} &  &  &  &  & \\
LiDAR (mode) & 0.61 & \textbf{0.58} & 1.25 & 1.98 & 3.33 \\
RAW (mean k\%) & 11.14 & \textbf{10.94} & 11.00 & 14.45 & 17.83 \\
RAW (mode) & 21.254 & \textbf{19.15} & 34.33 & 36.33 & 38.94 \\
REF (mean k\%) & 5.06 & 4.71 & 3.82 & 2.89 & \textbf{2.78} \\
REF (mode) & \textbf{1.33} & 1.34 & 2.00 & 2.77 & 4.12 \\
\bottomrule
\end{tabular*}
\end{table}

\item \noindent\textbf{Execution times:}
Figure~\ref{fig:time_result} presents the timing analysis over 2855 frames processed by the framework, excluding model loading and first-frame warm-up.
Producing only the MiDaS raw depth map required 440.35 ms. Adding the LiDAR point cloud, projected into the camera frame and augmented with simulated noise and dropout, increased processing to 855.05 ms. Moreover, refining MiDaS raw with the LiDAR sparse map further raised the processing time to 1703.18 ms, with timing distributions shifting rightward as each stage adds computational burden.

However, it is worth nothing that the current implementation prioritizes modularity and accuracy over efficiency: a high-capacity detector (YOLOv11x) and a high-accuracy MiDaS backbone (Swin2L-384) were selected, and the pipeline is implemented in Python for flexibility and reproducibility. No hardware- or inference-level optimizations (e.g., TensorRT, pruning/quantization, pipeline parallelization, or hardware co-processing) were applied so these results represent a baseline with room for real-time improvements.
Some overhead comes from point cloud reading, projection, and noise/dropout simulation; in real scenarios, pre-existing LiDAR data would reduce this burden.

\end{itemize}
\begin{figure}
    \centering
    \includegraphics[width=0.80\linewidth]{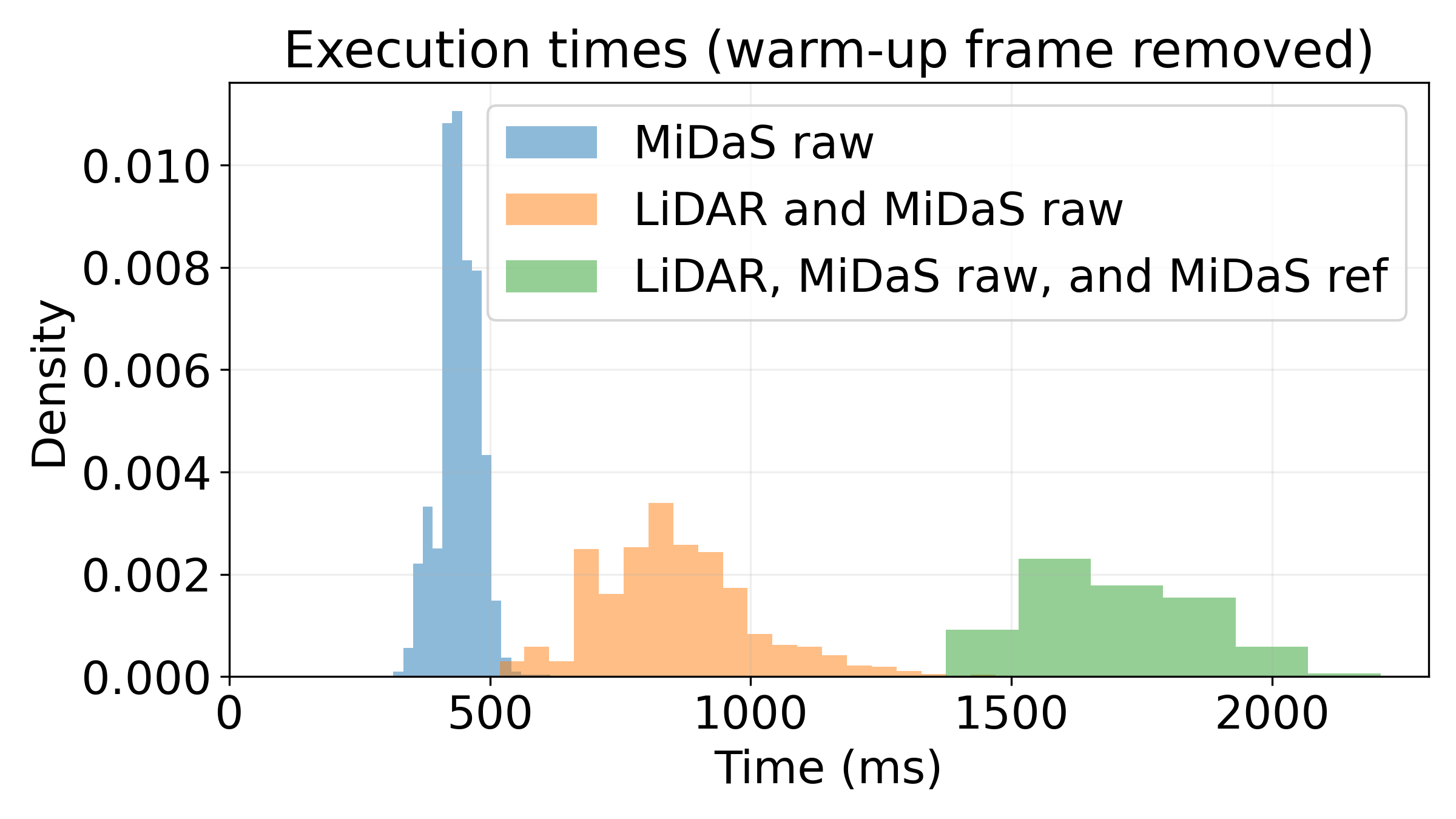}
    \caption{Distribution of execution times (warm-up frame removed) for three setups: MiDaS raw depth map only; LiDAR points and MiDaS raw depth map processed separately; and LiDAR points, MiDaS raw depth map, and MiDaS refined depth map outputs.}
    \label{fig:time_result}
\end{figure}

\section{Conclusions and future work}
\label{s:conclusions}

    This work presented a modular and flexible multi-sensor obstacle detection framework for railway environments, integrating track identification, object detection, and distance estimation through dedicated fusion algorithms. 
    
    The rail track segmentation network detects the railway track and its surrounding area, enabling the system to assess potential obstacles based on their position; the object detection network identifies and classifies only the object classes relevant to the railway environment, accurately localizing them through segmentation masks; and the monocular depth estimation network provides absolute depth estimates within object masks, excluding background outliers and improving generalization for railway scenarios. Moreover, the framework integrates distance information by fusing dense but less accurate monocular depth estimates with sparse, highly accurate measurements from the simulated LiDAR sensor.
    This produces a scene representation that is both spatially dense and metrically reliable, improving perception quality and obstacle distance estimation. In addition, tracking and filtering improve temporal consistency and distance prediction, further increasing the robustness of obstacle perception. Another important aspect of this work concerns the evaluation protocol. Given the scarcity of annotated railway datasets, the use of a synthetic dataset is not only a practical choice but also a methodological contribution. Experimental results demonstrate the feasibility of the proposed approach and confirm its effectiveness in enhancing scene understanding and supporting collision avoidance in railway applications.
    
    Nevertheless, some limitations remain. The system cannot detect objects outside the COCO classes and struggles with small or distant objects, a common issue for most object detection models that reflects an intrinsic limitation of such approaches. Moreover, the system has been developed and evaluated using synthetic data; as a result, a performance shift can be expected when deploying the framework on real-world data, which should be addressed in future evaluations. The framework is also not yet optimized for real-time performance, and further work is required to improve computational efficiency for time-critical operational contexts. Although the three networks execute in parallel, blocking still occurs because the object detection network must wait for the segmentation and monocular depth estimation outputs before integrating and annotating them in the final image. In addition, the data fusion block used to refine the MiDaS raw depth map improves distance perception but increases processing time.  
    
    Future work will focus on optimizing computational performance, systematically assessing the impact of each model within the modular framework, and further consolidating the proposed benchmarking methodology. Additional investigations will address robustness under diverse environmental conditions and explore more advanced fusion strategies to enhance the reliability and practical applicability of railway obstacle detection systems.

\section*{Acknowledgment}
The authors would like to thank Progress Rail Signaling S.p.A. for their support and collaboration in this project.
\bibliographystyle{IEEEtran}
\bibliography{bibliography}


 




\vfill

\end{document}